\documentclass[letterpaper, 10 pt, conference]{ieeeconf}  

\IEEEoverridecommandlockouts                              

\usepackage{graphicx}
\usepackage{hyperref}
\usepackage{caption}
\usepackage{subcaption}
\usepackage{multirow}
\usepackage[T1]{fontenc}
\usepackage{array}
\usepackage{float}
\usepackage{textcomp}
\usepackage{amsmath}
\usepackage{amssymb}
\usepackage{stackrel}
\usepackage{cite} 
\title{\LARGE \bf
Scalable Tactile Sensing for an Omni-adaptive Soft Robot Finger*
}
\author{Zeyi Yang$^{1,\#}$, Sheng Ge$^{1,\#}$, Fang Wan$^{2}$, Yujia Liu$^{1}$, and Chaoyang Song$^{3,*}$
\thanks{*This work was supported by Southern University of Science and Technology and AncoraSpring Inc.}
\thanks{$^{1}$Zeyi Yang,  Sheng Ge, and Yujia Liu are with Department of Mechanical and Energy Engineering, Southern University of Science and Technology, 
        Shenzhen, Guangdong 518055, China. 
        {\tt\small {11610518, 11612122, liuyj}@mail.sustech.edu.cn}}%
\thanks{$^{2}$Fang Wan is with AncoraSpring, Inc. and currently a visiting scholar at SUSTech Institute of Robotics, Southern University of Science and Technology, 
        Shenzhen, Guangdong 518055, China. 
        {\tt\small sophie.fwan@gmail.cn}}%
\thanks{$^{3}$Chaoyang Song is the corresponding author with the Department of Mechanical and Energy Engineering, Southern University of Science and Technology,
        Shenzhen, Guangdong 518055, China.
        {\tt\small songcy@ieee.org}}%
}
\begin{document}
\maketitle
\thispagestyle{empty}
\pagestyle{empty}
\begin{abstract}
    Robotic fingers made of soft material and compliant structures usually lead to superior adaptation when interacting with the unstructured physical environment. In this paper, we present an embedded sensing solution using optical fibers for an omni-adaptive soft robotic finger with exceptional adaptation in all directions. In particular, we managed to insert a pair of optical fibers inside the finger's structural cavity without interfering with its adaptive performance. The resultant integration is scalable as a versatile, low-cost, and moisture-proof solution for physically safe human-robot interaction. In addition, we experimented with our finger design for an object sorting task and identified sectional diameters of 94\% objects within the $\pm$6mm error and measured 80\% of the structural strains within $\pm$0.1mm/mm error. The proposed sensor design opens many doors in future applications of soft robotics for scalable and adaptive physical interactions in the unstructured environment. 
\end{abstract}
\section{Introduction}
\label{sec:Introduction}

    \begin{figure}[tbp]
        \begin{centering}
            \textsf{\includegraphics[width=1\columnwidth]{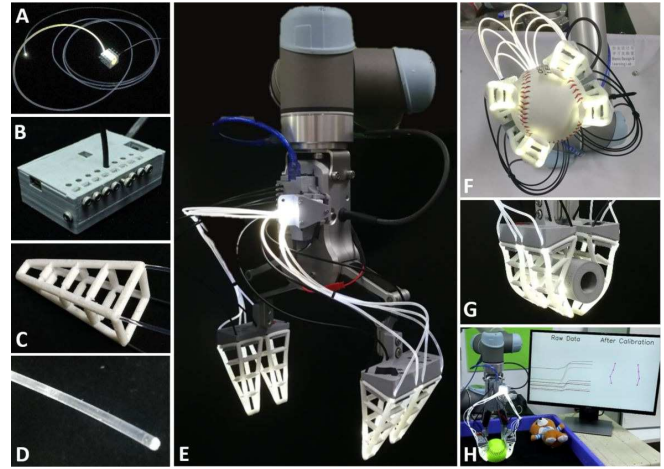}}
        \par\end{centering}
        \caption{Overview of scalable sensing design of gripper and sensors. The light source is a pair of strong LEDs installed in a box (A) that support eight optical fibers to be inserted. The photoresistance, microprocessor (Arduino NANO) and LED power are integrated into a small box (B) that can insert eight receiving fibers and connect a serial port as a power supply and communication interface. The grid-structured omni-adaptive soft finger (C) is simple and low-cost, which can be replaced easily. The material of the transparent optical fiber (E) is polymethyl methacrylate which has a 0.2-0.5db/m attenuation rate. The sensored fingertip is installed on a real robot arm UR5 (E). Each finger will passively adapt to a baseball (F) while grasping, especially be effective for cylindrical objects (G). The final integration of sensor with gripper can achieve detecting the horizontal section in real time (H). }
        \label{fig:PaperOverview}
    \end{figure}
    
    Robotic devices made of soft components not only exhibit superior adaptation in actuation \cite{Laschieaah3690}, but also in sensing \cite{wang2018toward, Polygerinos2017}. Previous research on tactile sensing usually requires explicit understanding of the material mechanics to build analytical models that translate structural deformation into sensory data \cite{Thuruthel2019, Wang2017}. However, the non-linear mechanics inherently involved in the soft material remains a challenging issue in the kinematic analysis and dynamic modeling of soft robot \cite{Wang2017, Tan2017, Martins2006, Rus2015, Boonvisut2012}. On the other hand, recent research has shown novel tactile sensing solution using soft robots through the integration with other devices, such as visual sensors \cite{Xu2019, Yuan2017}.

    The geometric response of the soft material provides a versatile source of information that captures the underlying dynamics during physical interaction \cite{Thuruthel2019, Truby2019, Hao2016}. The current development of visual sensors provides a robust mechanism for capturing the geometric deformations of the soft matter\cite{Wang2019}. 

    However, modeling the deformation of a soft structure is challenging. Numerous analyses and calculations were implemented to simulate a simple soft structure \cite{Tan2017}. The traditional sensing technology to evaluate the strain is using strain gauge which takes advantage of the properties of electrical principle \cite{Silva2002}, which requires the integration of the gauge in the fingers. Piezoresistance is a scalable and low-cost sensory element to generate tactile perception, but it also needs electrical arrays to support and is difficult to measure the bending state directly \cite{Chin2019, Zhao2016}. In addition, waterproofing must be considered, in case we use the gripper in a wet environment or even underwater, which is more challenging during tactile sensing integration.

\subsection{Related Work}
    Recent research about tactile sensor for soft robots focuses on innovations relating to scalability and engineering potentials \cite{Park2018}. The tendency of using neural networks to process high-dimensional sensory data is widely accepted by researchers, which provide a more accurate model via repetitious training \cite{Thuruthel2019, Yuan2017, Sundaram2019, Xu2019, Meerbeek2018}. However, the traditional calibration method is also applied by some sensors and shows good results \cite{Zhao2016, Chin2019}. Piezoresistive is an appropriate component for tactile sensing on soft robot \cite{Stassi2014}, its application on a perceptive glove shows the scalability on tactile sorting \cite{Sundaram2019}. A handmade capacitive stack-up sensor was tested and applied on an object sorting experiment \cite{Chin2019}, which is directly attached to a finger of handed shearing auxetic cylinders \cite{Lipton2018}. For soft robots, some structure allows us to embed sensors into their bodies and fuse them together, polydimethylsiloxane impregnated with conductive carbon nanotubes can be used as a strain sensor and it is small enough to be embedded into a soft structure as an integration \cite{Thuruthel2019}. 

    Optical sensing has been widely researched for its ease of integration with soft robots. An innovative method to detect the deformation of soft prosthetic hand via stretchable optical waveguides shows the prospect of an optical sensor \cite{Zhao2016}. A plastic optical fiber pressure sensor \cite{sartiano2017low} was presented as the merits of low cost and simple fabrication. And recent research about applying soft optoelectronic sensory foams presented an extremely accurate estimated 3D-model for entire deformation of a normal soft foam \cite{Meerbeek2018}.  A most recent research using optical lace also opens a window for soft robot tactile sensing \cite{Xu2019} using the contact of input fiber and distributed output fibers which are inserted in a 3D-printed elastomer. 

    Application is always a final goal of grippers and sensors. The soft grippers as the base of the tactile sensor have a variety of designs with diverse functions \cite{Shintake2018}. An embedded tactile sensor enables more functions for the gripper, such as closed-loop object picking \cite{Truby2019}. Besides, sorting experiment is a good verification for the properties of robotic soft fingers with tactile sensing. The application of fingers with the structure of handed shearing auxetic showed good examples of object sorting and material classification \cite{Chin2019, Chin2019b}.

\subsection{Proposed Method and Contributions}
    In this paper, we propose a scalable, embedded tactile sensing solution using soft plastic optical fiber inside a novel design of soft robot finger with passive, omni-directional adaptation, as shown in Fig. \ref{fig:PaperOverview}. While most tactile sensing solutions are usually considered as a subsystem independent to the overall robot, our proposed design is seamlessly integrated inside this unique network structure of the soft robotic finger without impeding its omni-adaptive performance. We managed to capture the three-dimensional geometric deformation through a scalable sensing solution using soft optical fiber. Major contributions of this paper are listed as the following.
    \begin{itemize}
        \item An integrated design of the fiber-cavity sensor with omni-adaptive soft finger.
        \item Extensive experiment and characterization of the fiber-cavity senor.
        \item Sensor implementation in sorting task of daily objects via an integrated gripper system.
    \end{itemize}

    The rest of this paper is structured as the following. Section \ref{sec:ScalableDesign} presents the design of the omni-adaptive finger network and the proposed tactile sensing solution. Section \ref{sec:SensChar} includes the experimental characterization of the omni-adaptive soft finger design using the the proposed tactile sensors. A demonstrative example is presented in section \ref{sec:ObjSort} to explore the usefulness of the proposed integrative design in object sorting tasks. Discussions and final remarks are enclosed in section \ref{sec:FinalRemarks}, which ends this paper.
\section{Embedded Tactile Sensing for Scalability}
\label{sec:ScalableDesign}
    \begin{figure}[bp]
        \begin{centering}
        \textsf{\includegraphics[width=1\columnwidth]{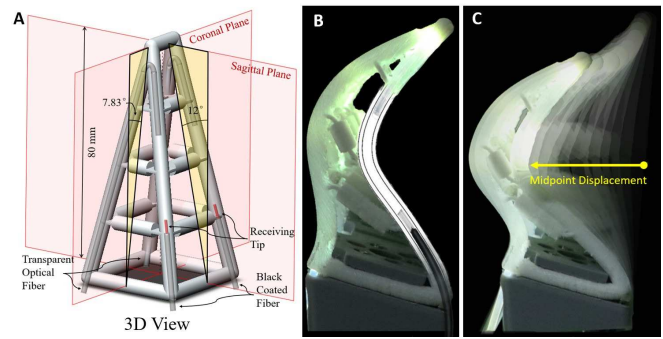}}
        \par\end{centering}
        \caption{Detailed design for the soft fingers and sensors. The front-view of the Omni-adaptive fingers are trapeziums. The side-view is a triangle. The Sagittal plane and Coronal plane are the main bending plane (A). The inside soft fiber passes through the inner tube to transmit the light which will be received by a black coated optical fiber (A). The tube wall will hinder light transmission while bending (B). The midpoint is the most crucial point while grasping so that can be considered as the standard value to measure the degree of bending (C).}
        \label{fig:OmniFingerDesign} 
    \end{figure}

\subsection{Soft Finger Network for Omni-adaptation}
     In this paper, we adopt a novel design of soft finger network with passive adaptation in all directions of physical contact. Fig. \ref{fig:OmniFingerDesign}A shows the three-dimensional (3D) view of the soft finger network, where layers of squared shapes with shrinking area are stacked on top of the other with links on the sides to connect them, forming the basic structure of this finger design. When fabricated with soft material, such as silicone rubber or Thermoplastic Urethane (TPU), the 3D structure is capable of passive adaptation of the overall structural geometry, as shown in Figs. \ref{fig:OmniFingerDesign}B \& C. Due to the hollowed squares used, the finger achieves omni-directional adaptation instead of a uni-directional response. In fact, one can design any shape for each of the layers as long as certain hollow can be kept near the certain of each layer for geometric adaptation. 

\subsection{Embedded Optical Fiber for Scalable Tactile Sensing}
    Given the omni-adaptive nature of this soft finger network, we set our sensor design with a goal of minimum interference with its geometric adaptation without limiting its usage scenario. As a result, the optical fibers are selected for several reasons. First, the material property of the optical fibers is very similar to that of the some materials used for this soft finger network. Second, optical sensing is capable of robust measurement over a long distance and the optical sensor is actually not placed in the finger structure, but outside of it near the gripper base. As a result, we can still apply such soft finger design in the same operational environment without worrying about the protection of the sensing electronics on the finger. Finally, optical fibers is a relatively cheap solution when scalability is taken into considerations.

    We implement the resultant sensor design by creating a cavity with the structural supporting beams between each finger layers, and then embedding the optical fibers inside to capture the geometric deformation. The transmitting fiber are different with the receiving fiber. The core of transmitting optical fiber (Model hof-2, EverHeng Optical Co., Shenzhen, China) is 2mm polymethyl methacrylate (PMMA) fiber, and with a cladding of transparent polytetra fluoroethylene (PTFE) outside. The receiving fiber (Model epef-1.5) is 1.5mm PMMA core with PTFE cladding and additional black polyvinyl chloride (PVC) jacket. For example, in the soft finger structure with four supporting beams shown in Fig. \ref{fig:OmniFingerDesign}, cylindrical cavity is designed inside each of the beams matching the diameter of the optical fiber. From the side-viewing angle in Fig. \ref{fig:OmniFingerDesign}B, the transmitting optical fiber with a light source is inserted from the base of one beam at the back side of interaction. Then, the receiving optical fiber is inserted through another beam at the front side of interaction all the way to its base, where photo-resistance sensors (Model GL5506) are installed. During bending motions, the reading from the sensors correspond to the amount of geometric deformations inside the front side beams of interaction. When the backbones bend in a direction, the received light intensity will attenuate theoretically because of being hindered by the deformation of the tube wall. We named the sensor as a fiber-cavity sensor. In addition, the length of the inner cavity needs to be carefully selected. After several testing, 35mm shows a satisfied result of performance. Too long or too short will cause the reduction of the measurement range.

    The resultant design achieves tactile sensing through an experimental mapping between the geometric deformation of the soft finger network and the differential readings from the optical fibers, which correspond to various geometric features of the objects during physical interactions. We achieve a rich set of readings when multiple sets of such fiber-cavity sensors are used. The differences between different sensor sets provides more detailed information of the geometric deformation in 3D.

\section{Sensor Characterization}
\label{sec:SensChar}
\begin{figure}[tbp]
        \begin{centering}
            \textsf{\includegraphics[width=1\columnwidth]{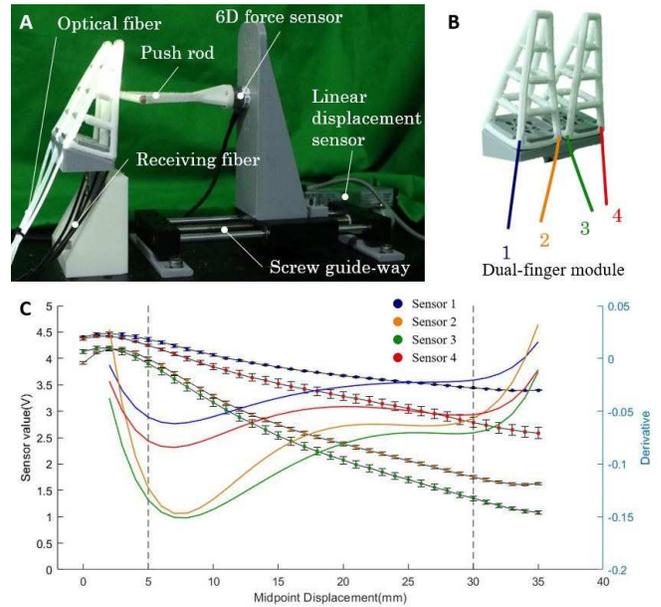}}
        \par\end{centering}
        \caption{Experimental platform (A) and data from fiber-cavity sensors. The half of gripper is a dual-finger model with four-channel of fiber-cavity sensor (B), which was directly used in the experiment. The graph is about the raw data of sensors (line with square and error bar) and its derivative (thinner fine line) after filtering (C). The same color represents the same sensor. }
        \label{fig:ExpResult} 
    \end{figure}

    Given the nature of our integrated sensor design, the experiment setup is closely related to observations of the soft finger network under loading. Beside the pure bending behaviour at the normal surface, the omni-directional adaptation relies greatly on the twisting deformation at random angles to the finger surface, which is essentially a differential readings from the two contacting beams. As a result, we setup our experiment characterization by measuring the force normal to the finger surface as shown in Fig. \ref{fig:ExpResult}A. A T-shape rod is fixed to a mount on top of a manual linear guide-way to push the midpoint of the finger at right angle as the displacement input. A 6D force and torque sensor (ATI Nano 17) is fixed at the end of the T-shape rod for measuring the output force. Optical sensor readings are also recorded for sensor characterization and calibration. 

    In this experiment, two such soft finger networks are mounted at the same time, which is the same as the ones to be installed on one of the finger tips of the robotic gripper to be used later. A total of four sets of fiber-cavity sensor readings are recorded in Fig. \ref{fig:ExpResult}B with results reported in Fig. \ref{fig:ExpResult}C. We definite the original point of displacement at the midpoint of the front backbone in the non-grasping state. The positive direction is towards the back surface. The sensor value is a voltage from 0v to 5v, which positively correlates with the light intensity. Each measurement is repeated three times and the standard error bars are also included. 

    We identify three stages of behaviours from the results in Fig. \ref{fig:ExpResult}C between a measurement range of 0-35mm displacement range. For the initial stage up to around 5mm displacement at finger midpoint, the small bending behaviours of the soft finger network is not well-captured by the fiber-cavity sensors. The diffuse reflection of the light by the tube wall causes the photo-transduction instability, invalidating sensor readings at this stage. For the final stage beyond 30mm displacement at finger midpoint, although the soft finger network still shows adaptive behaviours, the inner layers starts to stack on top of each other as shown in Fig. \ref{fig:OmniFingerDesign}C, making it difficult to produce consistent sensor readings. Sensors readings during this stage is also disregarded by one can still utilize the twisting behaviour at this stage for grasping object of irregular shape.

    \begin{figure}[tbp]
        \begin{centering}
            \textsf{\includegraphics[width=1\columnwidth]{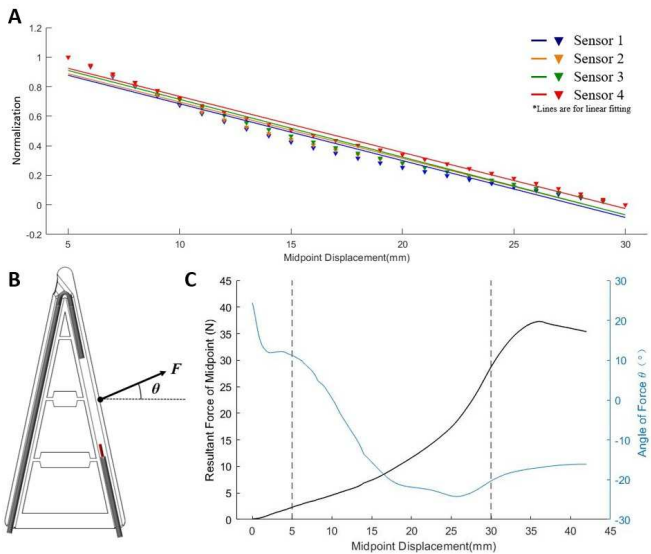}}
        \par\end{centering}
        \caption{Linear fitting in the interval of 5mm to 30mm (A). Magnitude and angle of the contact force at midpoint (B) change with displacement (C). The graph is just for one finger, if using a dual-finger model, the magnitude will become twice as much as the current curve.}
        \label{fig:LinearResult}
    \end{figure}

    During the stable stage between 5-30mm displacement at finger midpoint, the recorded results shows good linearity between displacement and sensor readings in voltage changes, making this stage the most suitable for usage. The results in Fig. \ref{fig:ExpResult}C shows slightly differences in the sensors placed at the same locations on the two soft finger network, but consistent results are recorded. We found that this is caused by the fabrication errors and assembly inaccuracies, which can be improved with optimized engineering processing and sensor calibration shown in Fig. \ref{fig:LinearResult}A. After normalization and linear fitting, the sensor can be regarded as a linear element that relates to the midpoint displacement in Fig. \ref{fig:LinearResult}A. The $R^2$ value is within 0.9544 and 0.9887, which is acceptable for linear fitting. Therefore, this interval of the curve can be regarded as a linear variation that can be used for sensor integration.

    Tactile sensing information is extracted by mapping the displacement readings from the optical sensors with the force measurement from the 6D FT sensors, as shown in Fig. \ref{fig:LinearResult}C. The measured displacement-force relationship shows consistent results after long-hours of usage and the reliable linear performance from previous experiments. We found the data was basically the same as the origin for any finger after our calibration and sorting experiments during two weeks. Therefore, by using the results in Figs. \ref{fig:ExpResult}C and Fig. \ref{fig:LinearResult}C, one can derive the force information of the soft finger structure during interaction. Alternatively, one can also detect the hardness to measure the strain under a constant force. Basically, The softer the object is, the greater the strain will be. 
\section{Object Sorting}
\label{sec:ObjSort}
    In this section, we aim at establishing the tactile sensing potentials of the fiber-cavity sensor for object sorting via collaborative robot Universal Robot UR5. Being a passive finger with omni-adaptive capability, the soft finger structure provides an enabling functionality to exiting grippers of rigid design with shape adaptation. As shown in Fig. \ref{fig:PaperOverview}, we propose a dual-finger design where two of such finger structures are mounted on a simple and small flange to replace the exiting gripper's common rigid finger. In the following experiment setup, the OnRobot RG6 is adopted for modification with two sets of the dual-finger structure. The RG6 is selected for its relatively large range of grasping and heavy payload design. One can easily modify the base mount design according to different fingers to install this proposed soft finger structures on almost any robotic grippers with rigid finger structure, introducing scalable and enabling capability for shape adaptation in grasping tasks. 

    For the object sorting task, both YCB objects \cite{Calli2015} and some other routine objects are chosen for experiments. A microcomputer is integrated with the fiber-cavity sensors on the gripper to send all sensor values to the upper computer by a serial port in real-time. The grasping force of RG6 needs to be set by users, but RG6 will measure and feedback the current width of the gripper. The additional sensing capability introduced by the fiber-cavity sensor enables further refined control of the grasping process by estimating the interaction force and shape geometry of the objects in contact. The actual sectional diameter of the object under certain force theoretically equals to the estimated midpoint displacement plus the width of RG6 at the beginning contact point. And the actual strain of the object can be calculated by measuring the width of the object at both beginning contact state and final steady-state.  

\subsection{Calibration}
\begin{figure}[htbp]
        \begin{centering}
            \textsf{\includegraphics[width=1\columnwidth]{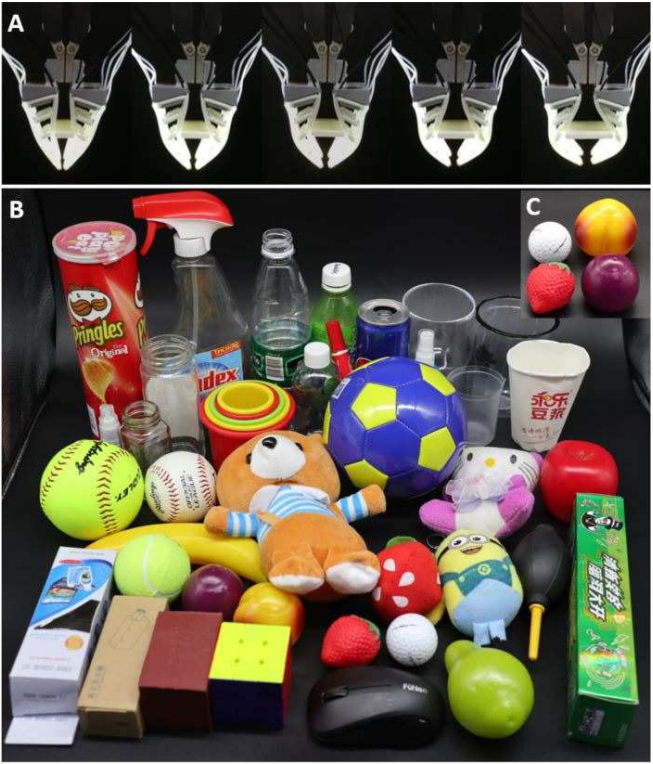}}
        \par\end{centering}
        \caption{Fast calibration for gripper in a real application (A). Plates with different plates will be placed between the gripper and record the sensor value for further linear fitting. A selection of sorting objects are selected from the YCB dataset and daily objects (B). The basic idea of selection is to ensure the the group of sectional diameter and compliance of the object is unique. Due to the parallel finger configuration used in our experiment, some objects (C) may slip through the gap between dual-fingers during grasping.}
        \label{fig:CalibSetup}
    \end{figure}

    Calibration of the sensor needs to transfers the raw measurement of the electronics data into intuitive information of physical values. To do so, we use a series of plates with different standard widths to calibrate the gripper (Fig. \ref{fig:CalibSetup}A). Although the resolution of senor is less than 0.2mm, the inaccuracy is beyond this range. Therefore, we implement the calibration process by letting the gripper grasp the plates with standard width and record the sensor value. After obtaining a group of data, linear fitting will be used to obtain a proportional relationship as the calibrated result of the sensor. So, any sensor readings will be transferred to the midpoint displacement via the calibrated expression. 

\subsection{Sorting Experiment}
    Some objects with different sizes and compliance were selected in the sorting experiment in Fig. \ref{fig:CalibSetup}B. One obvious challenge to distinguish objects with sectional diameters similar to each other. However, this is not common in the YCB object sets used in our experiment. We adopt a qualitative measurement of the object hardness in a way similar to human grasping, where a scalar level of hardness is adopted. It should be noted that more accurate measurement is always preferable, yet different grasping compliance may occur when approached from different angles. So the strains of samples under a certain force were manually determined by observing and simple measuring. The standard strains of samples are not absolutely accurate but in accordance with the common sense of human, which can be used to judge the estimated strain. For object classification, our experiment requires the gripper to squeeze the object to determine this geometric features for sorting, which is similar to human when visual data is not available or sufficiently enough. In this way, if the force applied to the object is constant, the strain of the object will be different due to the different compliant characteristics. The strain of each object is the ratio of deformation and original width before being exerted a certain force. The ratio should be different in different materials, which can be used to distinguish the objects (Fig. \ref{fig:GraspPred}).

\subsection{Result}
\begin{figure}[htbp]
        \begin{centering}
            \textsf{\includegraphics[width=1\columnwidth]{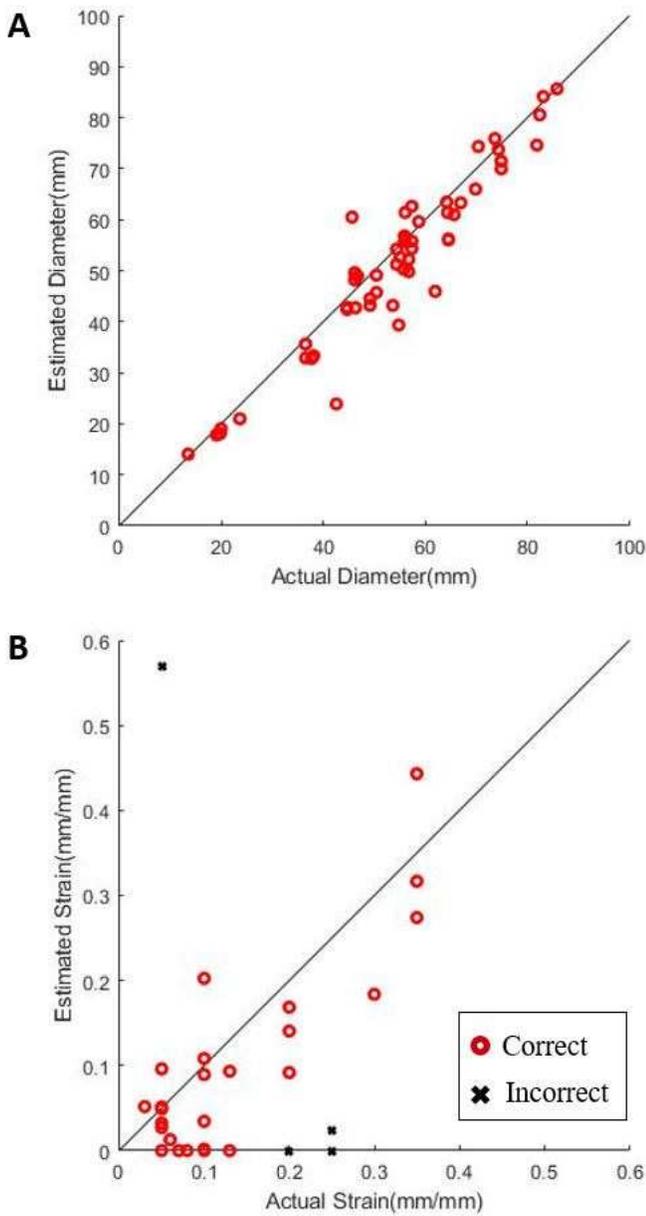}}
        \par\end{centering}
        \caption{Comparison of actual object properties with senor estimated properties. The estimated sectional diameter showed a nearly linear relationship with the actual diameter (A). 94\% of rigid objects are within the error of $\pm$6mm, expect the abnormal objects. The estimated strain, which represents the compliance of the object, has a relative consistency with the actual strain (B). 80\% of 28 objects, which strain can be measured, are within the error of $\pm$0.1mm/mm.The accuracy of strain estimation is not perfect, but the relative compliance rank is basically correct. The cross line is the soft-rigid boundary. The upper-right part is soft objects and the lower-left part is rigid objects.}
        \label{fig:GraspPred} 
    \end{figure}

    The results are reported in Fig. \ref{fig:GraspPred}, where we explore the basic discernibility of the gripper for width and compliance. The total amount of sample objects is 42. The green triangular marks as shown in Fig. \ref{fig:GraspPred}A are 9 softest objects whose estimated diameters are much smaller than the actual diameters because their structures or materials cannot support the finger force. The orange diamond marks are the 2 balls whose diameters adapt the finger space but will cause the lateral bending and torsion of the finger. The lateral bending will result in the underestimate of the diameter in sagittal plane. The black square mark is the result of container of glass cleaner. The overestimate error happened because once a pair of fingers contact the bigger diameter of the bottle, it will prevent the other pair of fingers to contact smaller part. Thus, the one result is normal but the other is abnormal. 8 objects cannot be measured their strain because the midpoint displacement did not reach the valid interval from 5mm to 30mm. The total amount of objects whose diameter and compliance can be correctly measured is 28 in 38, so the success rate of object classification is 73.7\%.

    Expect the soft objects, balls and irregular object, 94\% of results from the rest 26 objects are in the range of $\pm$6mm with respect to absolutely accurate as shown in Fig. \ref{fig:GraspPred}A. The average error of the estimated diameter is 3.17mm. The result comes from the sectional diameter measured by two pairs of fiber-cavity sensors. There have four pairs of sensors, but we only use the two intermediate pairs of sensors to ensure the fingers entirely contact the objects, and the finger is passively driven by the shape of the object.  

    Twenty-eight objects has appropriate data to estimate strain from total graspable 38 objects. The result was shown in the Fig. \ref{fig:GraspPred}B. We cannot get a accurate conclusion as the actual strains are intuitive perception of human, but a qualitative analysis is possible. The black dashed cross line is the boundary of rigid and soft according to the experiment. The objects in upper-right section are deformable, such as plush toys. The objects in the lower-left section are rigid. The average error of the estimated strain is 0.062mm/mm. Some objects cannot induce enough deformation of the finger, which cannot calculate the strain via sensor data. So, for those objects, we consider them as unrecognizable samples. The estimated strain can be used to describe the hardness of an object because the grasping force is always the same. With the increasing of strain, the objects become softer. 
\section{Discussion}
\label{sec:Discussion}
\subsection{Scalable Integration of Omni-adaptive Soft Finger}
    The scalability of the fiber-cavity sensory gripper is its greatest merit. First, the whole strategy is simple and low-cost that total cost of the four-fingers gripper with eight fiber-cavity sensors and one microcomputer (Fig. \ref{fig:PaperOverview}E) is less than 8 US dollars and the time of assembling all sensors into one 3D-printed finger is less than one minute. Thus, the modular sensory fingers could be as daily using or even more short-term using. Second, the structure could be used in many aspects, not only in the grasping area, and with the changing of the whole shape, sensor strategy can easily adapt to the new shape without modification. For example, it can be used as a wheel to adapt the topography or an exoskeleton to adapt the wearer's body. Third, current fingers do not need to embed circuits, so working in a wet environment is its additional merit. Integrating these merits above, one of the most suitable working cases is in waste sorting. We do not need to consider the water in garbage and sterilization and disinfection method for the finger. Forth, the soft material and flexible structure could enable any rigid gripper a kind of omni-adaptability, and the scalable fiber-cavity sensor enable it a tactile sense. At last, the fiber-cavity sensor still has great potentiality because of high distinguishability and sensitivity after calibrating.  

\subsection{Enabling Design for Omni-adaptation}
    A major advantage of such soft finger network is ease of integration with existing gripper designs. By replacing the finger tips with this soft finger network, almost any rigid gripper is instantly enabled with passive adaptation with superior performance in all contact directions, such as the one shown in Fig. \ref{fig:PaperOverview}. Further discussion of this finger design is beyond the scope of this paper. In this paper, we aim at utilizing such geometric adaptation to integrate a sensing solution within the finger network structure. 

\subsection{Engineering Application for Object Sorting}
    Our current fiber-cavity sensory gripper is still in its early stage, which has some aspects that should be improved. First, the ambient light has some effect on the sensor value as the material is not lightproof. Although the calibrating operations could eliminate the influence of ambient light, the change of light after calibration still has some impact on the sensor. The receiving optical fiber is black coated, so the transparency of the white TPU cavity's wall mainly caused the sensitivity to ambient light. Second, the inconsistency of the fiber-cavity sensor is a problem, even the normalizing process could unify the curves, but that is not a permanent solution. The inconsistency is mainly caused by the fabricating process, which completely made by hand. In addition, the 3D-printed TPU finger has some defects and burrs on the inner tube wall that also affect the light transmission. Third, to pursue extremely low-cost, high sensitivity in week light environment and wide sensing range, we apply photoresistors as our underlying sensory elements. However, the inconsistency of the element, temperature-dependent and non-linear property are its drawback. 

    The current drawbacks of the fiber-cavity sensor are not insoluble. For the sensitivity of ambient light, a grey or black colored TPU material could reduce transmissivity widely or just coat a light-absorption layer to avoid ambient light influence. For the inconsistency of the fiber-cavity sensor, we are trying to use casting to manufacture the finger whose surface is smooth and fine. In addition, we will try to use some photoresistors with higher accuracy and inconsistency or find an appropriate photodiode. 

    The result of estimating diameter (Fig. \ref{fig:GraspPred}A) is significant for further work, because the obvious regulation for the soft object shows the potential of fusion of visual and tactile sensing. The visual diameter can be fast measured by a extra camera but the material is unknowable. With the fiber-cavity sensor, we can take advantage of the variation of diameter after grasping to estimate hardness as the softer objects will have more difference between visual size and tactile size. 

    As for the result of estimating strain (Fig. \ref{fig:GraspPred}A), the quantitative conclusion is inaccurate. There are three possible reasons. First, the standard strains of specific objects are measured by human perception. We ensured the rank of each object's strain was correct but were unable to guarantee that the absolute value was correct. So, the distribution of points is scattered, but tendency of points basically concentrate at a correct area. The conclusion of 'soft' or 'rigid' for a certain object is correct according to the value of sensors and the soft-rigid boundary. Second, we think the sensor is accurate in the range of 5-30mm, but some objects cannot reach 5mm of midpoint displacement, because their structural shape limit the bending of fingers, whose contact points are at the tips of fingers. Third, the compliance of 3D-printed finger are inconsistent and the sectional diameter of object is variable, both of which result in the heterogeneous force while grasping. So, the result of estimating strain for the same objects by different pairs of sensors is differentiated. But for human perception, the hardness of object is a scale rather than a numerical value, so we consider the result is useful as the tendency is correct.
\section{Final Remarks}
\label{sec:FinalRemarks}
    In this paper, we demonstrated a scalable bending sensor method for a novel design of omni-adaptive soft robotic fingers. Our work combined the omni-adaptive finger and fiber-cavity sensor to enable more functions based on its own structure. And we implemented experiments to find the relationship between midpoint displacement and sensor value and demonstrated their nearly linear relationship in the interval of 5mm to 30mm, which could be used to calibrate the gripper and estimate the actual width and compliance of objects in sorting tasks. The final result of sorting showed the estimated widths of 94\% objects are within $\pm$6mm error and the estimate strains of 80\% objects are within $\pm$0.1mm/mm. The object identification rate from a total of 38 objects from the YCB dataset and some other objects covering basic routine things is 73.7\%. 

    Future work on this sensory gripper will focus on the improvement of the fiber-cavity sensor and application in multi-tasking. We will continue to develop the advantages of low-cost and modular design, meanwhile, to improve the performance of the fiber-cavity sensor. Our next application scenarios are for waste sorting, which needs the properties of omni-adaptive, waterproof, low-cost, and easy-replaced. We would like to combine the computer vision and neural network to complete a more perceptive model of the finger and build a system to learn how to grasp new objects through training.
                                  
\bibliographystyle{IEEEtran}
\bibliography{references}

\begin{thebibliography}{10}
\providecommand{\url}[1]{#1}
\csname url@rmstyle\endcsname
\providecommand{\newblock}{\relax}
\providecommand{\bibinfo}[2]{#2}
\providecommand\BIBentrySTDinterwordspacing{\spaceskip=0pt\relax}
\providecommand\BIBentryALTinterwordstretchfactor{4}
\providecommand\BIBentryALTinterwordspacing{\spaceskip=\fontdimen2\font plus
\BIBentryALTinterwordstretchfactor\fontdimen3\font minus
  \fontdimen4\font\relax}
\providecommand\BIBforeignlanguage[2]{{%
\expandafter\ifx\csname l@#1\endcsname\relax
\typeout{** WARNING: IEEEtran.bst: No hyphenation pattern has been}%
\typeout{** loaded for the language `#1'. Using the pattern for}%
\typeout{** the default language instead.}%
\else
\language=\csname l@#1\endcsname
\fi
#2}}

\bibitem{Laschieaah3690}
\BIBentryALTinterwordspacing
C.~Laschi, B.~Mazzolai, and M.~Cianchetti, ``Soft robotics: Technologies and
  systems pushing the boundaries of robot abilities,'' \emph{Science Robotics},
  vol.~1, no.~1, 2016. [Online]. Available:
  \url{https://robotics.sciencemag.org/content/1/1/eaah3690}
\BIBentrySTDinterwordspacing

\bibitem{wang2018toward}
H.~Wang, M.~Totaro, and L.~Beccai, ``Toward perceptive soft robots: Progress
  and challenges,'' \emph{Advanced Science}, vol.~5, no.~9, p. 1800541, 2018.

\bibitem{Polygerinos2017}
P.~Polygerinos, N.~Correll, S.~A. Morin, B.~Mosadegh, C.~D. Onal, K.~Petersen,
  M.~Cianchetti, M.~T. Tolley, and R.~F. Shepherd, ``Soft robotics: Review of
  fluid-driven intrinsically soft devices; manufacturing, sensing, control, and
  applications in human-robot interaction,'' \emph{Advanced Engineering
  Materials}, vol.~19, no.~12, p. 1700016, 2017.

\bibitem{Thuruthel2019}
T.~G. Thuruthel, B.~Shih, C.~Laschi, and M.~T. Tolley, ``Soft robot perception
  using embedded soft sensors and recurrent neural networks,'' \emph{Science
  Robotics}, vol.~4, no.~26, p. eaav1488, 2019.

\bibitem{Wang2017}
Z.~Wang and S.~Hirai, ``Soft gripper dynamics using a line-segment model with
  an optimization-based parameter identification method,'' \emph{IEEE Robotics
  and Automation Letters}, vol.~2, no.~2, pp. 624--631, 2017.

\bibitem{Tan2017}
N.~Tan, X.~Gu, and H.~Ren, ``Simultaneous robot-world, sensor-tip, and
  kinematics calibration of an underactuated robotic hand with soft fingers,''
  \emph{IEEE Access}, vol.~6, pp. 22\,705--22\,715, 2017.

\bibitem{Martins2006}
P.~Martins, R.~Natal~Jorge, and A.~Ferreira, ``A comparative study of several
  material models for prediction of hyperelastic properties: Application to
  silicone-rubber and soft tissues,'' \emph{Strain}, vol.~42, no.~3, pp.
  135--147, 2006.

\bibitem{Rus2015}
D.~Rus and M.~T. Tolley, ``Design, fabrication and control of soft robots,''
  \emph{Nature}, vol. 521, no. 7553, p. 467, 2015.

\bibitem{Boonvisut2012}
P.~Boonvisut and M.~C. {\c{C}}avu{\c{s}}o{\u{g}}lu, ``Estimation of soft tissue
  mechanical parameters from robotic manipulation data,'' \emph{IEEE/ASME
  Transactions on Mechatronics}, vol.~18, no.~5, pp. 1602--1611, 2012.

\bibitem{Xu2019}
P.~A. Xu, A.~Mishra, H.~Bai, C.~Aubin, L.~Zullo, and R.~Shepherd, ``Optical
  lace for synthetic afferent neural networks,'' \emph{Science Robotics},
  vol.~4, no.~34, p. eaaw6304, 2019.

\bibitem{Yuan2017}
W.~Yuan, C.~Zhu, A.~Owens, M.~A. Srinivasan, and E.~H. Adelson,
  ``Shape-independent hardness estimation using deep learning and a gelsight
  tactile sensor,'' in \emph{2017 IEEE International Conference on Robotics and
  Automation (ICRA)}.\hskip 1em plus 0.5em minus 0.4em\relax IEEE, 2017, pp.
  951--958.

\bibitem{Truby2019}
R.~L. Truby, R.~K. Katzschmann, J.~A. Lewis, and D.~Rus, ``Soft robotic fingers
  with embedded ionogel sensors and discrete actuation modes for
  somatosensitive manipulation,'' in \emph{2019 2nd IEEE International
  Conference on Soft Robotics (RoboSoft)}.\hskip 1em plus 0.5em minus
  0.4em\relax IEEE, 2019, pp. 322--329.

\bibitem{Hao2016}
Y.~Hao, Z.~Gong, Z.~Xie, S.~Guan, X.~Yang, Z.~Ren, T.~Wang, and L.~Wen,
  ``Universal soft pneumatic robotic gripper with variable effective length,''
  in \emph{2016 35th Chinese Control Conference (CCC)}.\hskip 1em plus 0.5em
  minus 0.4em\relax IEEE, 2016, pp. 6109--6114.

\bibitem{Wang2019}
R.~Wang, S.~Wang, E.~Xiao, K.~Jindal, W.~Yuan, and C.~Feng, ``Real-time soft
  robot 3d proprioception via deep vision-based sensing,'' \emph{arXiv preprint
  arXiv:1904.03820}, 2019.

\bibitem{Silva2002}
J.~G. da~Silva, A.~A. de~Carvalho, and D.~D. da~Silva, ``A strain gauge tactile
  sensor for finger-mounted applications,'' \emph{IEEE Transactions on
  Instrumentation and measurement}, vol.~51, no.~1, pp. 18--22, 2002.

\bibitem{Chin2019}
L.~Chin, M.~C. Yuen, J.~Lipton, L.~H. Trueba, R.~Kramer-Bottiglio, and D.~Rus,
  ``A simple electric soft robotic gripper with high-deformation haptic
  feedback,'' in \emph{International Conference on Robotics and Automation},
  2019.

\bibitem{Zhao2016}
H.~Zhao, K.~O'Brien, S.~Li, and R.~F. Shepherd, ``Optoelectronically innervated
  soft prosthetic hand via stretchable optical waveguides,'' \emph{Science
  Robotics}, vol.~1, no.~1, p. eaai7529, 2016.

\bibitem{Park2018}
M.~Park, B.-G. Bok, J.-H. Ahn, and M.-S. Kim, ``Recent advances in tactile
  sensing technology,'' \emph{Micromachines}, vol.~9, no.~7, p. 321, 2018.

\bibitem{Sundaram2019}
S.~Sundaram, P.~Kellnhofer, Y.~Li, J.-Y. Zhu, A.~Torralba, and W.~Matusik,
  ``Learning the signatures of the human grasp using a scalable tactile
  glove,'' \emph{Nature}, vol. 569, no. 7758, p. 698, 2019.

\bibitem{Meerbeek2018}
I.~Van~Meerbeek, C.~De~Sa, and R.~Shepherd, ``Soft optoelectronic sensory foams
  with proprioception,'' \emph{Science Robotics}, vol.~3, no.~24, p. eaau2489,
  2018.

\bibitem{Stassi2014}
S.~Stassi, V.~Cauda, G.~Canavese, and C.~F. Pirri, ``Flexible tactile sensing
  based on piezoresistive composites: A review,'' \emph{Sensors}, vol.~14,
  no.~3, pp. 5296--5332, 2014.

\bibitem{Lipton2018}
J.~I. Lipton, R.~MacCurdy, Z.~Manchester, L.~Chin, D.~Cellucci, and D.~Rus,
  ``Handedness in shearing auxetics creates rigid and compliant structures,''
  \emph{Science}, vol. 360, no. 6389, pp. 632--635, 2018.

\bibitem{sartiano2017low}
D.~Sartiano and S.~Sales, ``Low cost plastic optical fiber pressure sensor
  embedded in mattress for vital signal monitoring,'' \emph{Sensors}, vol.~17,
  no.~12, p. 2900, 2017.

\bibitem{Shintake2018}
J.~Shintake, V.~Cacucciolo, D.~Floreano, and H.~Shea, ``Soft robotic
  grippers,'' \emph{Advanced Materials}, vol.~30, no.~29, p. 1707035, 2018.

\bibitem{Chin2019b}
L.~Chin, J.~Lipton, M.~C. Yuen, R.~Kramer-Bottiglio, and D.~Rus, ``Automated
  recycling separation enabled by soft robotic material classification,'' in
  \emph{2019 2nd IEEE International Conference on Soft Robotics
  (RoboSoft)}.\hskip 1em plus 0.5em minus 0.4em\relax IEEE, 2019, pp. 102--107.

\bibitem{Calli2015}
B.~Calli, A.~Singh, A.~Walsman, S.~Srinivasa, P.~Abbeel, and A.~M. Dollar,
  ``The ycb object and model set: Towards common benchmarks for manipulation
  research,'' in \emph{2015 international conference on advanced robotics
  (ICAR)}.\hskip 1em plus 0.5em minus 0.4em\relax IEEE, 2015, pp. 510--517.

\end{thebibliography}
\end{document}